\documentclass[runningheads]{llncs}
\usepackage{graphicx}
\usepackage{verbatim}
\usepackage{multirow}
\usepackage{booktabs}
\usepackage{array}
\usepackage{authblk}

\renewcommand{\thefootnote}{\fnsymbol{footnote}}

\begin{document}
\title{Pre-Training with Diffusion models for Dental Radiography segmentation 
%\thanks{Supported by organization Allisone Technologies}
}

\titlerunning{PTDR}

\author{
    Jérémy Rousseau \and
    Christian Alaka$^*$  \and
    Emma Covili$^*$ \and 
    Hippolyte Mayard$^*$ \and
    Laura Misrachi$^*$ \and 
    Willy Au 
}

\authorrunning{J. Rousseau et al}

\institute{
    Allisone Technologies, Paris, France \url{https://www.allisone.ai/}\\
    \email{\{jeremy, christian, emma, hippolyte, laura, willy\}@allisone.ai} 
}

\maketitle % typeset the header of the contribution
\def\thefootnote{*}\footnotetext{These authors contributed equally to this work}\def\thefootnote{\arabic{footnote}}
\begin{abstract}
Medical radiography segmentation, and specifically dental radiography, is highly limited by the cost of labeling which requires specific expertise and labor-intensive annotations. 
In this work, we propose a straightforward pre-training method for semantic segmentation leveraging Denoising Diffusion Probabilistic Models (DDPM), which have shown impressive results for generative modeling.
Our straightforward approach achieves remarkable performance in terms of label efficiency and does not require architectural modifications between pre-training and downstream tasks.
We propose to first pre-train a Unet by exploiting the DDPM training objective, and then fine-tune the resulting model on a segmentation task. Our experimental results on the segmentation of dental radiographs demonstrate that the proposed method is competitive with state-of-the-art pre-training methods.

\end{abstract}

\section{Introduction}

Accurate automatic semantic segmentation of radiographs is of high interest in the dental field as it has the potential to help practitioners identify anatomical and pathological elements more quickly and precisely. 
While deep learning methods show robust performances at segmentation tasks, they require a substantial amount of pixel-level annotations which is time-consuming and demands strong expertise in the medical field. 
Accordingly, many recent state-of-the-art methods \cite{MoCo,MoCo2,simCLR,simMIM,BEIT,SWIN_MAE}  use  self-supervised learning as a pre-training step to improve training and reduce labeling effort in computer vision.

Inspired by the renewed interest in denoising for generative modeling, we investigate denoising as a pre-training task for semantic segmentation. 
Denoising autoencoder is a classic concept in machine learning where a model learns to separate the original data from the noise, and implicitly learns the data distribution by doing so \cite{Vincent2008,Vincent2010}. 
In particular, denoising objective can be easily defined pixel-wise, making it especially well suited for segmentation tasks~\cite{DDeP}.

Recently, a new class of generative models, known as Denoising Diffusion Probabilistic Models (DDPM) \cite{DDPM,sohl2015deep,ImprovedDiffusion}, have shown impressive results for generative modeling. DDPM outperform other state-of-the-art generative models such as Generative Adversarial Networks (GANs)~\cite{GAN} in various tasks, including image synthesis \cite{DiffusionBeatsGAN}.
 
DDPM learn to convert Gaussian noise to a target distribution via a sequence of iterative denoising steps, yielding impressive results in image synthesis outperforming GANs \cite{DiffusionBeatsGAN,GAN}.

\begin{figure}[!ht]
    \centering
    \includegraphics[width=\textwidth]{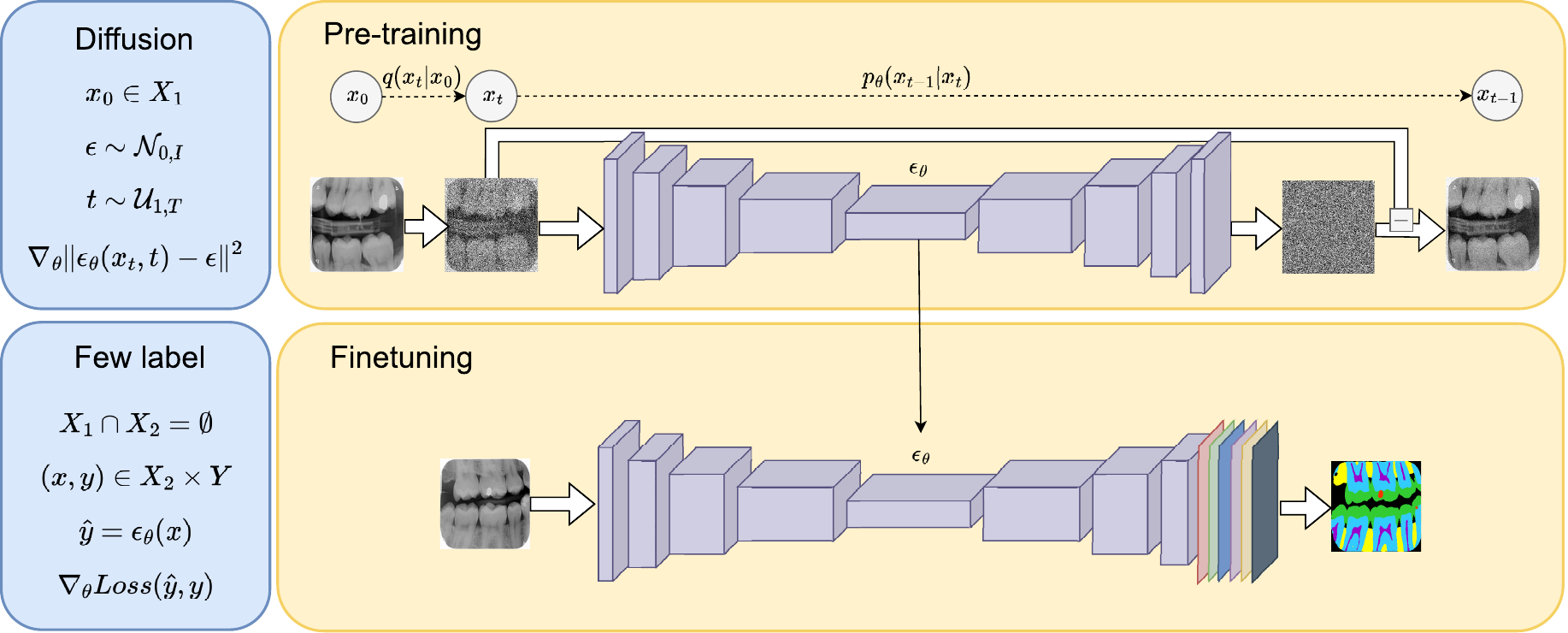}
    \caption{\textbf{PTDR method overview.}\\ \textit{top} - $\mathbf{\epsilon}_\theta$ is pre-trained on unlabeled dataset $X_1$ using the training procedure of DDPM \cite{DDPM}. \textit{bottom} - $\mathbf{\epsilon}_\theta$ is then fine-tuned on a small labeled dataset $X_2$. $Y$ represents the set of ground truth semantic maps.}
    \label{fig:archi}
\end{figure}

Following the success of DDPM for generative modeling, \cite{SegDiff,DDPM_seg_ensemble,MedSegDiff,MedSegDiffv2} explore their ability to directly generate semantic maps in an iterative process by conditioning each denoising steps with a raw image prior. ~\cite{Yandex} shows that DDPM are effective representation learners whose feature maps can be used for semantic segmentation, beating previous pre-training methods in a few label regime.

In this paper, we propose Pre-Training with Diffusion models for
Dental Radiography segmentation (PTDR). The method consists in pre-training a Unet~\cite{Unet} in a self-supervised manner by exploiting the DDPM training objective, and then fine-tuning the resulting model on a semantic segmentation task.

To sum up our contributions, our method is most similar to \cite{Yandex} but does not require fine-tuning a different model after pre-training. 
The whole Unet architecture is pre-trained in one step at the difference of~\cite{DDeP} which requires two.
At inference, only one forward pass is used, making it easier to use than \cite{Yandex,SegDiff}.
Finally, we show that our proposed method surpasses other state-of-the-art pre-training methods especially when only few annotated samples are available.

\section{Methodology}

    \subsection{Background} \label{part:2.1}

Inspired by Langevin dynamics, DDPM~\cite{DDPM} formalize the generation task as a denoising problem where an image is 
gradually corrupted for $T$ steps and then reconstructed through a learned reverse process. 
Generation is done by applying the reverse process to pure random noise. \\

Starting from an image $\mathbf x_{0}$, the forward diffusion process iteratively produces noisy versions of the image $\mathbf{\{{x}_{t}\}_{t=1}^{T}}$, and is defined as a Gaussian Markov chain where $\mathbf{\{\beta_{t}\in \left(0,1\right)\}_{t=1}^{T}}$ is the variance schedule: 
\begin{equation}
    q\left(\mathbf{x}_t \mid \mathbf{x}_{t-1}\right):=\mathcal{N}\left(\mathbf{x}_t ; \sqrt{1-\beta_t} \mathbf{x}_{t-1}, \beta_t \mathbf{I}\right)
\end{equation}

A noisy image $\mathbf{x}_t$ is obtained at any timestep $\mathbf{t}$ from the original image $\mathbf{x}_0$ with the following closed form, let $\mathbf{\alpha _{t}=1-\beta_{t}}$ and $\bar{\alpha}_t=\prod_{s=1}^t \alpha_s$ we have:
\begin{equation}
q\left(\mathbf{x}_t \mid \mathbf{x}_0\right)=\mathcal{N}\left(\mathbf{x}_t ; \sqrt{\bar{\alpha}_t} \mathbf{x}_0,\left(1-\bar{\alpha}_t\right) \mathbf{I}\right)
\end{equation}

When the diffusion steps are small enough, the reverse process can also be modeled as a Gaussian Markov chain:

\begin{equation}
p_\theta\left(\mathbf{x}_{t-1} \mid \mathbf{x}_t\right):=\mathcal{N}\left(\mathbf{x}_{t-1} ; \mathbf{{\mu}_\theta}\left(\mathbf{x}_t, t\right), {\sigma}_t ^{2} \mathbf{I}\right)
\end{equation}
\newline
where:
\begin{equation}\mu_\theta\left(\mathbf{x}_t, t\right)= \frac{1}{\sqrt{\alpha_t}}\left(\mathbf{x}_t-\frac{1-\alpha_t}{\sqrt{1-\bar{\alpha}_t}} 
\mathbf{\epsilon}_\theta\left(\mathbf{x}_t, t\right)\right) \quad {\sigma}_t ^{2} = \frac{1-\bar{\alpha}_{t-1}}{1-\bar{\alpha}_t} \beta_t
\end{equation}
with $\mathbf{\epsilon}_\theta$ the neural network being optimized. \\

 The training procedure is finally derived by optimizing the usual variational bound on the negative log-likelihood, and consists of randomly drawing samples $\mathbf{\epsilon}\sim\mathcal{N}_{0,\mathbf{I}}$,  $\mathbf{t} \sim \mathcal{U}_{1,\mathbf{T}}$, $\mathbf {x}_0\sim q({x}_0)$ and taking a gradient step on 
\begin{equation}
    \nabla _{\theta}\left\| \epsilon_{\theta} \left(\sqrt{\overline{\alpha}_{t}}x_{0}+\sqrt{1-\overline{\alpha}_{t}} \epsilon,t\right) -\epsilon\right\|^{2}
\end{equation}

    \subsection{DDPM for semantic segmentation}
    
The proposed method is based on two steps. First, a denoising model is pre-trained on a large set of unlabeled data following the procedure presented in section~\ref{part:2.1}. Second, the model is fine-tuned for semantic segmentation on few annotated data of the same domain by minimizing the cross-entropy loss.

Our method is similar to~\cite{Yandex} which leverages a pre-trained DDPM-based model as a feature extractor. 
Their method involves upsampling feature maps from predetermined activation blocks - from several forward passes at different timesteps - to the target resolution and training an ensemble of pixel-wise classifiers on concatenated feature maps. 
\cite{Yandex} showed that semantic information carried by feature maps highly depends on the activation block and the diffusion timestep. The latter are thus important hyper-parameters that need to be tuned for each specific semantic task.
This method originally introduced in~\cite{DatasetGAN} - in the context of GANs - is well-suited for generative models feature extraction but does not leverage the DDPM architecture as PTDR does.  

Our approach, by simply re-using the DDPM-trained denoising model for the downstream task, does not need extra classifiers and does not depend on activation blocks hyper-parameter. 
Moreover, PTDR fine-tuning and inference phases only require one forward pass in which the timestep is fixed to a predetermined value. 
To that extent, the proposed method is simpler both in terms of training and inference.

\section{Experiments and Results}

    \subsection{Experimental Setup}

In our experiments, a Unet* \footnote{Unet* denotes the specific Unet architecture introduce
in~\cite{DiffusionBeatsGAN}} based DDPM is trained on unlabeled radiographs, the Unet* is then
fine-tuned on a multi-class semantic segmentation task as illustrated in figure~\ref{fig:archi}. 
We experiment with regimes of 1, 2, 5 and 10 training samples and compare our results to other state-of-the-art self-supervised pre-training methods.
We used a single NVIDIA T4 GPU for all our experiments.\\

\begin{figure}[!ht]
    \centering
    \includegraphics[width=0.8\textwidth]{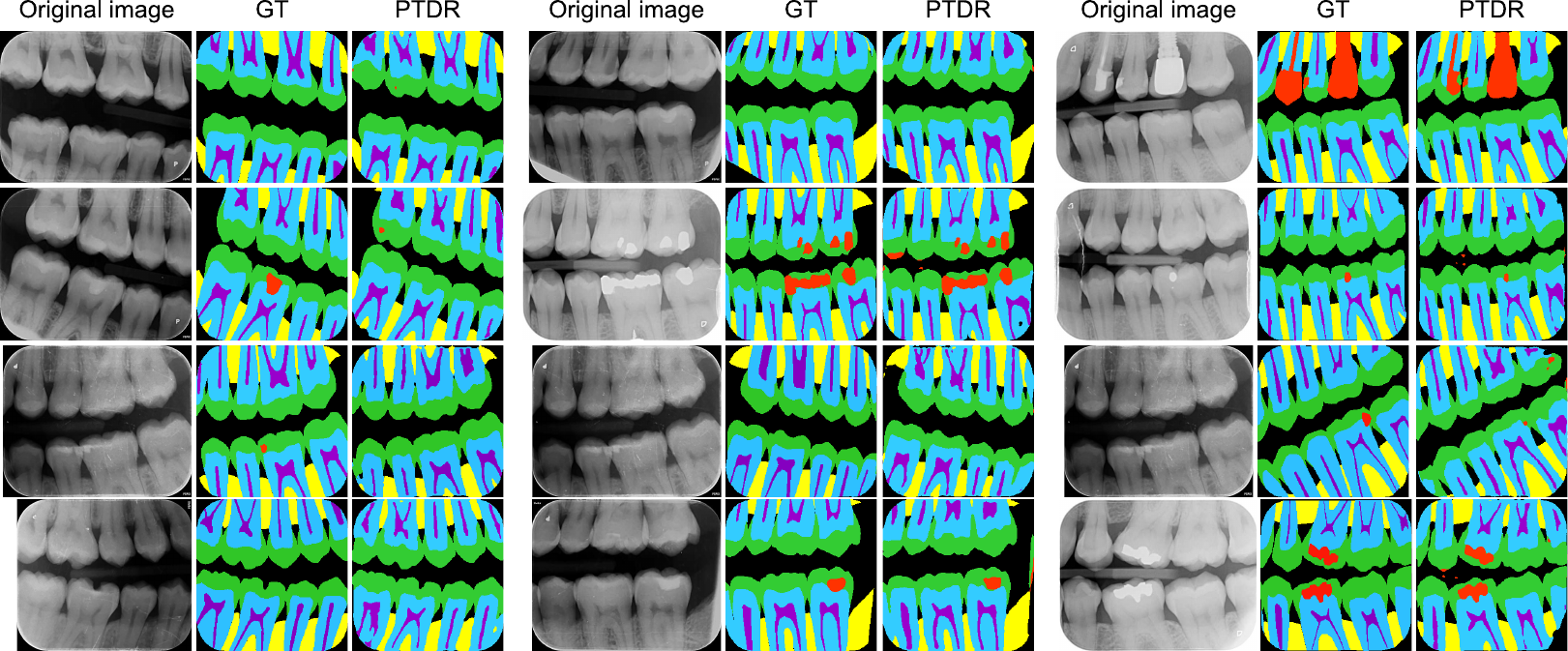}
    \caption{\textbf{Comparison} on test dental bitewing radiographs of ground truth (GT) against predicted semantic maps from PTDR fine-tuned on 10 labeled images.}
    \label{fig:multi-viz}
\end{figure}

\noindent\textbf{Datasets:} \label{part:datasets} Our main experiment is done on dental bitewing radiographs collected from partner dentists, see figure~\ref{fig:multi-viz}. The pre-training dataset contains 2500 unlabeled radiographs. Additionally, 100 bitewing radiographs are fully annotated for 6 classes namely: \textit{dentine}, \textit{enamel}, \textit{bone}, \textit{pulp}, \textit{other} and \textit{background} as semantic maps; and is randomly split into 10 training, 5 validation, and 85 test samples. There is no intersection between the pre-training and fine-tuning dataset. For our experiments, we use random subsets of the train set of size 1, 2, 5 and 10 respectively. 
Images are resized to 256x256 resolution and normalized between -1 and 1.\\

\noindent \textbf{Pre-training:}  The Unet* implemented in pytorch is trained with a batch size of 2 and follows the training procedure of~\cite{DiffusionBeatsGAN} with 4000 diffusion steps $T$. We use the official pytorch implementation of~\cite{DiffusionBeatsGAN}.
The training was performed for 150k iterations and we saved the weights at iteration 10k, 50k, 100k and 150k for fine-tuning comparison.\\

\noindent \textbf{Fine-tuning:} The batch size is set at 2. 
We use a random affine augmentation strategy with the following parameters: rotation angle uniformly sampled from $[-180, 180]$, shear sampled from $[-5, 5]$, scale sampled from $[0.9, 1.1]$, and translate factor sampled from $[0.05, 0.05]$. 
Fine-tuning is done for 200 epochs using the Adam  optimizer~\cite{Adam} with a learning rate of $1e^{-4}$, a weight decay of $1e^{-4}$, and a cosine scheduler.\\

\noindent \textbf{Baseline methods:}
The DDPM training procedure is performed for 150k iterations and used for both PTDR and~\cite{Yandex} which is referred to as DDPM-MLP for the next sections.
We also pre-train a Unet* encoder with MoCo v2~\cite{MoCo2} and then fine-tune the whole network on the downstream task. We refer to this method as MoCo v2.
Finally, we pre-train a Swin Transformer~\cite{SWIN} using SimMIM~\cite{simMIM} and use it as an Upernet~\cite{UperNet} backbone. We refer to this method as SimMIM. 
As the Swin backbone relies on batch normalization layers, we do not train SimMIM in the 1-shot regime. 
For all these methods, we use the same hyper-parameters as proposed in the original papers.\\

\noindent \textbf{Evaluation metric:} We use mean Intersection over Union (mIoU) as our evaluation metric to measure the performance of the downstream segmentation task.

    \subsection{Results}

We compare our method with other baseline pre-training methods and compare their performances on the multi-class segmentation downstream task in the 10-labeled regime as shown in table~\ref{tab:results}.

\begin{table}[!ht]
\centering
\caption{Comparison of pre-training methods when fine-tuned on 10 labeled samples}\label{tab:results} 
\begin{tabular}{ p{3cm} p{3cm} p{1cm} } \toprule
Model                           & Pre-training               & mIoU             \\ \midrule
\multirow{2}{*}{SwinUperNet}    & --                         & 59.58            \\ 
                                & SimMIM~\cite{simMIM}       & 70.69            \\ \midrule
\multirow{4}{*}{Unet*}          & --                         & 61.40            \\ 
                                & MoCo v2~\cite{MoCo2}       & 64.10            \\ 
                                & DDPM-MLP~\cite{Yandex}     & 69.64            \\ 
                                & \textbf{PTDR (ours)}       & \textbf{76.96}   \\ \bottomrule
\end{tabular}
\end{table}

\begin{figure}[!ht]
    \centering
    \includegraphics[width=0.75\textwidth]{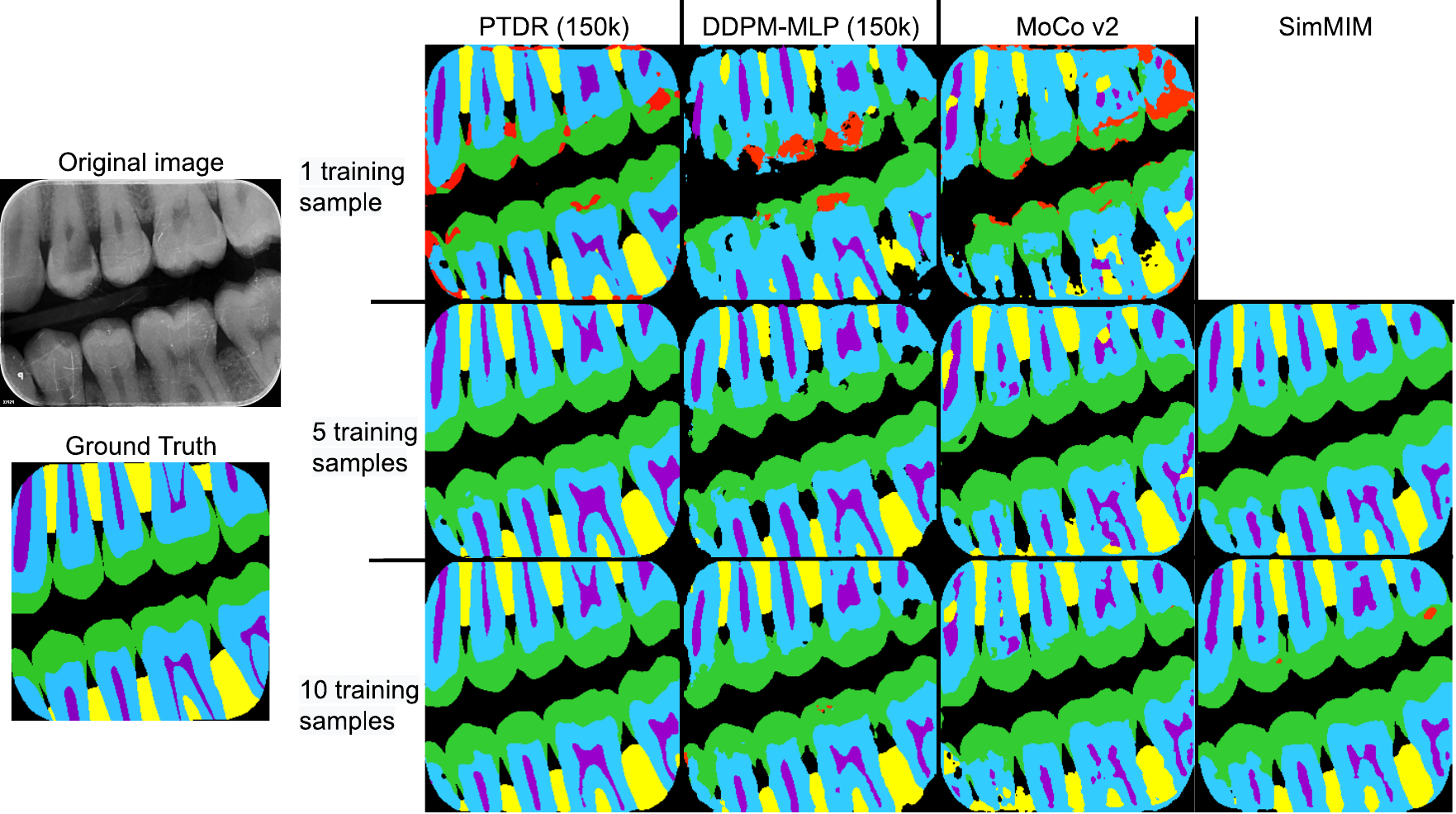}
    \caption{\textbf{Semantic maps} produced by different methods, PTDR, DDPM-MLP, MoCo v2 and SimMIM. The DDPM pre-training procedure is performed for 150k iterations. Semantic maps were produced by models trained on 1, 5, and 10 training samples to illustrate label efficiency.}
    \label{fig:visu}
\end{figure}

Our method outperforms all other methods, improving upon the second-best method by 10.5\%. Qualitative results on bitewing radiographs are shown in figure~\ref{fig:visu} with predicted semantic maps produced by all compared methods for 1, 5, and 10 training samples. For all regimes, predictions from our method are less coarse than others.

\begin{figure}[!ht]
    \centering
    \includegraphics[width=0.75\textwidth]{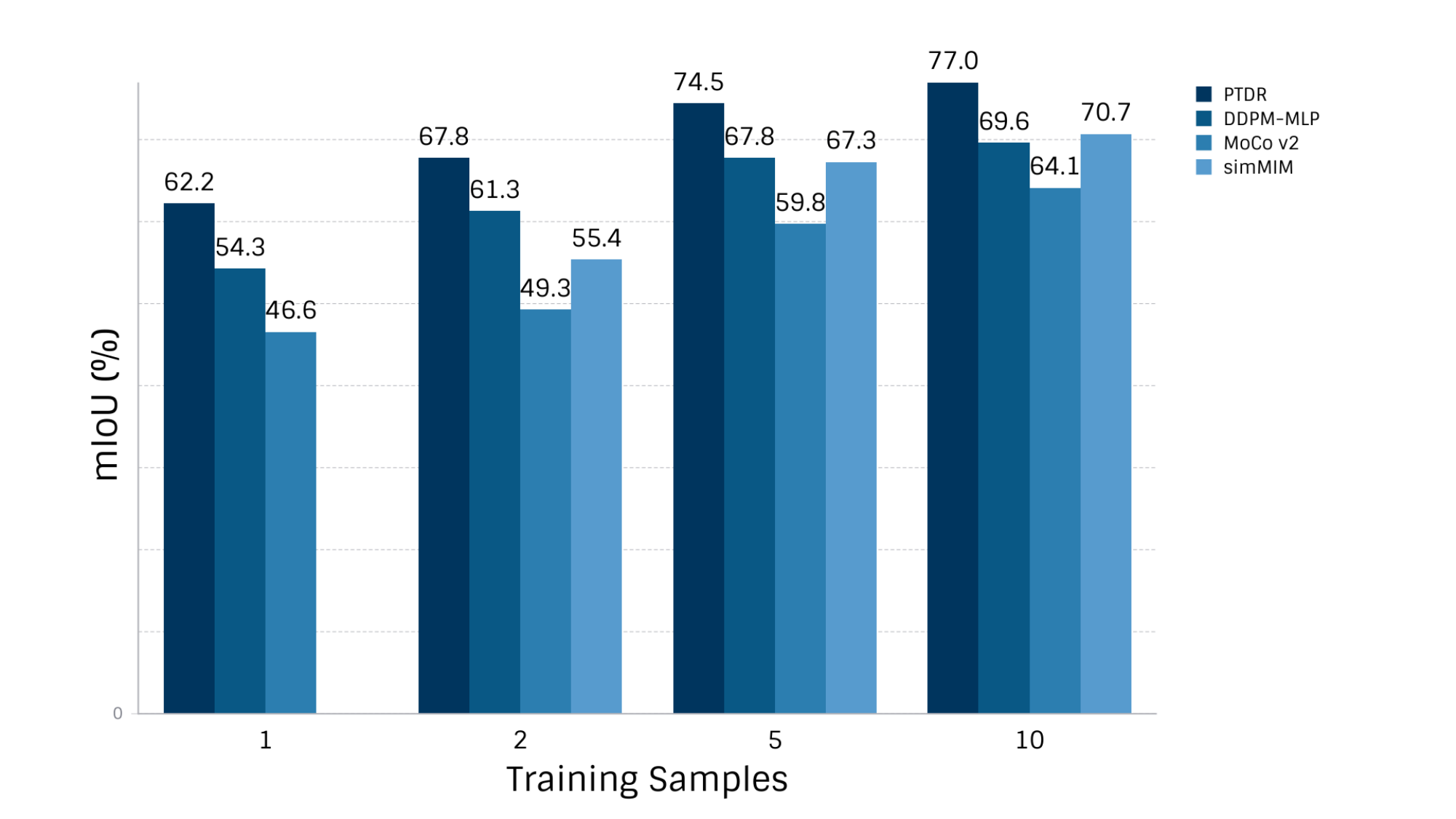}
    \caption{\textbf{Label efficiency.} Comparison of pre-training methods when fine-tuning in several data (1, 2, 5, and 10 training samples).}
    \label{fig:efficiency}
\end{figure}

\noindent \textbf{Label efficiency:} In this experiment, we compare our method with baseline methods in different data regimes.
Figure~\ref{fig:efficiency} illustrates the comparison between methods fine-tuned on 1, 2, 5, and 10 training samples.

Results show that our method yields better performance, in any regime, than all other pre-training methods benchmarked.
On average, over all regimes, PTDR improves upon DDPM-MLP, its closest competitor, by 7.08\%.
Moreover, we can observe in figure~\ref{fig:efficiency}, that our method trained on only 5 training samples outperforms all other methods trained on 10 samples. \\

\begin{figure}[!ht]
    \centering
    \includegraphics[width=\textwidth]{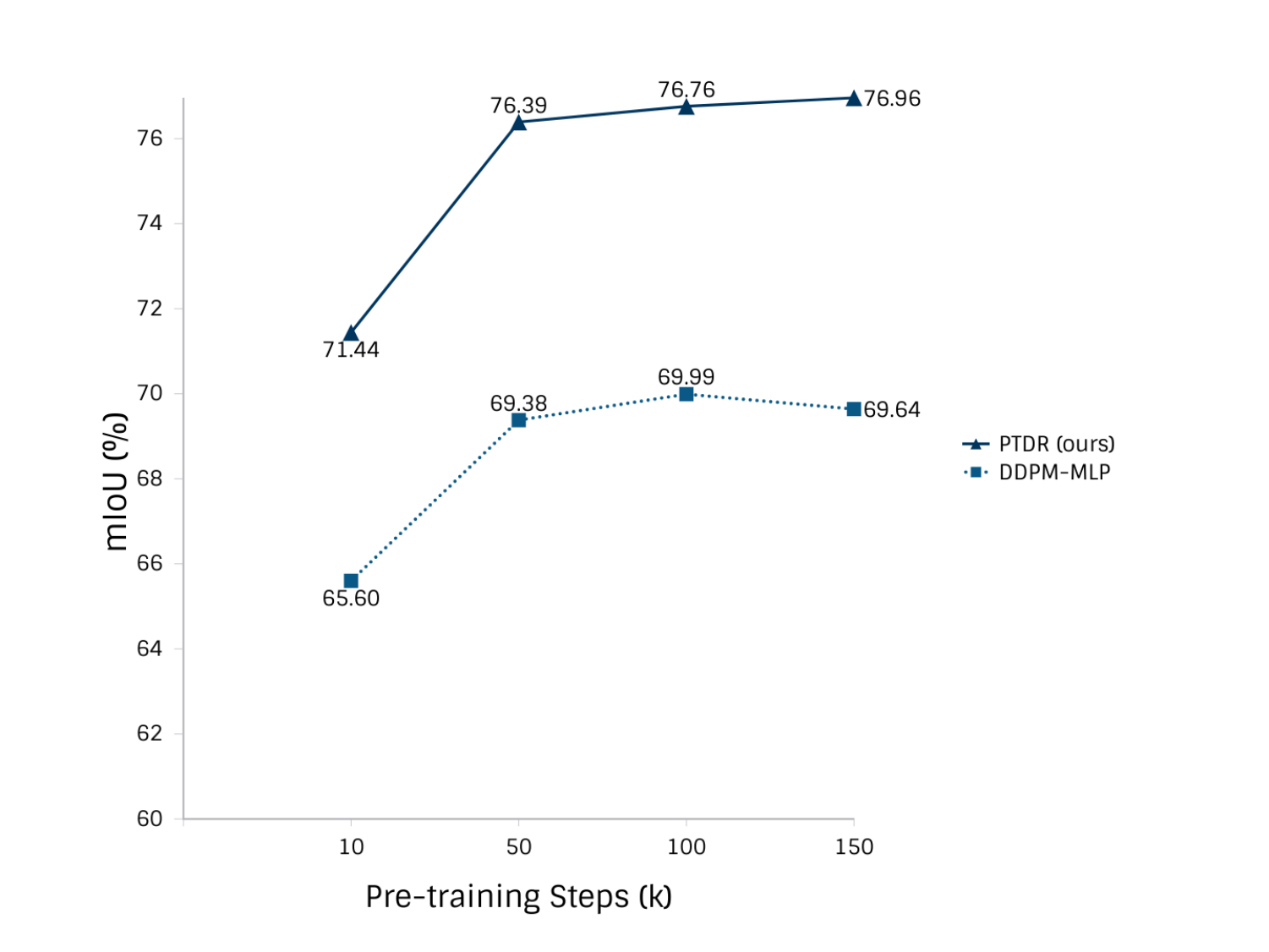}
    \caption{\textbf{Saturation effect.} Impact of the number of pre-training steps on mIoU for PTDR and DDPM-MLP trained on 10 training samples.}
    \label{fig:saturation}
\end{figure}

\noindent \textbf{Saturation effect:} We explore the influence of the number of DDPM pre-training iterations on the per final segmentation performance. In figure~\ref{fig:saturation}, we observe strong benefits of pre-training between 10k and 50k iterations with an absolute mIoU increase of +7\% for PTDR and +6\% for DDPM-MLP. As we advance in iteration steps, the pre-training effectiveness decreases. For both methods, we observe that beyond 50k iterations, the performance \textit{saturates} reaching a plateau. This suggests pre-training DDPM can be stopped before reaching ultra-realistic generative performance while still providing an efficient pre-trained model.
\\

\noindent \label{time_step_influence}\textbf{Timestep influence:} We investigate the influence of timestep, which conditions the Unet* and the amount of Gaussian noise added during the diffusion process. We empirically show in table~\ref{tab:timestep} that timestep 1 is the optimal setup during fine-tuning. This is intuitive as this timestep corresponds to the first diffusion step during which images are almost not corrupted which mirrors the fine-tuning setup on raw images. We did not find any benefits from letting the network learn the timestep value. However, it is worth mentioning that when we do so, the timestep converges to 1.\\

\begin{table}[!ht]
\centering
\caption{Influence of timestep value on PTDR's fine-tuning performance}\label{tab:timestep}
    \begin{tabular} {p{3cm} p{1.2cm} p{1.2cm} p{1.2cm} p{1.2cm} p{1.2cm} p{1.2cm}} \toprule
    Timestep value  & 1               & 100    & 1000   & 2000   & 4000   & learnt \\ \midrule
    mIoU            & \textbf{76.96}  & 76.94  & 76.61  & 74.86  & 73.60  & 76.80 \\ \bottomrule
    \end{tabular}
\end{table}

\textbf{Generalization capacity}: In appendix A, we further investigate the generalization capacity of our method to another medical dataset. \\

\textbf{Dataset generation}: In appendix B, we qualitatively illustrate the method's ability to generate a high-quality artificial dataset with pixel-wise labels.

\section{Conclusion}
This paper proposes a method that consists of two steps: a self-supervised pre-training using denoising diffusion models training objective and a fine-tuning of the obtained model on a radiograph semantic segmentation task. 
Experiments on dental bitewing radiographs showed that PTDR outperforms baseline self-supervised pre-training methods in the few label regime. 
Our simple, yet powerful, method allows the fine-tuning phase to easily exploit all the representations learned in the network during the diffusion pre-training phase without any architectural changes. 
These results highlight the effectiveness of diffusion models in learning representations. 
In future works, we will investigate the application of this method to other types of medical datasets.

\bibliography{bibliography}
\bibliographystyle{splncs04}

\newpage

\section*{Appendix A: Generalization Capacity}
We further tested PTDR on another multi-class semantic segmentation task of lung axial CT images to explore the capacity of our method to transfer to other modalities.
For this experiment, Radiopaedia Covid-19 dataset~\cite{DatasetGAN} (829 slices) is used for pre-training and COVID-19 CT Segmentation dataset~\cite{DatasetGAN} (100 slices) is used for fine-tuning. 
The latter is annotated for 4 classes: \textit{ground-glass}, \textit{consolidation}, \textit{lung-other} and \textit{background} as semantic maps; and is randomly split with 10 training, 5 validation and 85 test samples. 
The CT-slices are resized to 256x256, clipped between -1100 and +300 and normalized according to the mean and standard deviation of the clipped Radiopaedia dataset. 
There is no intersection between the pre-training and fine-tuning dataset. We show that the good performances of the proposed method are not restricted to bitewing radiographs and might be used for other types of medical image segmentation, as illustrated by results of table~\ref{tab:generalization_results}.

\begin{table}[!ht]
    \centering
    \caption{Performance of PTDR and DDPM-MLP on lung CT images segmentation}
    \label{tab:generalization_results}
    \begin{tabular}{ p{2cm} p{3cm} p{1cm} } \toprule
    Model                           & Pre-training               & mIoU             \\ \midrule
    \multirow{2}{*}{Unet*}          & DDPM-MLP~\cite{Yandex}     & 74.64           \\ 
                                    & \textbf{PTDR (ours)}       & \textbf{81.10}   \\ \bottomrule
    \end{tabular} 
\end{table}

\section*{Appendix B: Dataset Generation}
We study the ability of the proposed method to generate an artificial dataset. First, a set of dental radiograph is generated using DDPM then a semantic map is generated for each image using both PTDR and DDPM-MLP~\cite{Yandex}. We demonstrate qualitatively that the examples generated by PTDR are more consistent than those generated by DDPM-MLP. This capacity paves the way for application such as transfer learning or digital clones.

\begin{figure}[!ht]
    \centering
    \includegraphics[width=\textwidth]{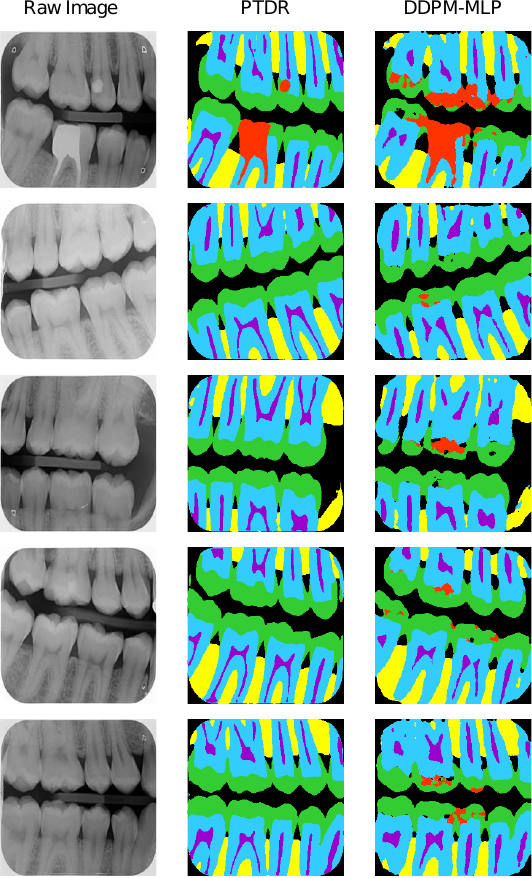}
\end{figure}

\begin{figure}[!ht]
    \centering
    \includegraphics[width=\textwidth]{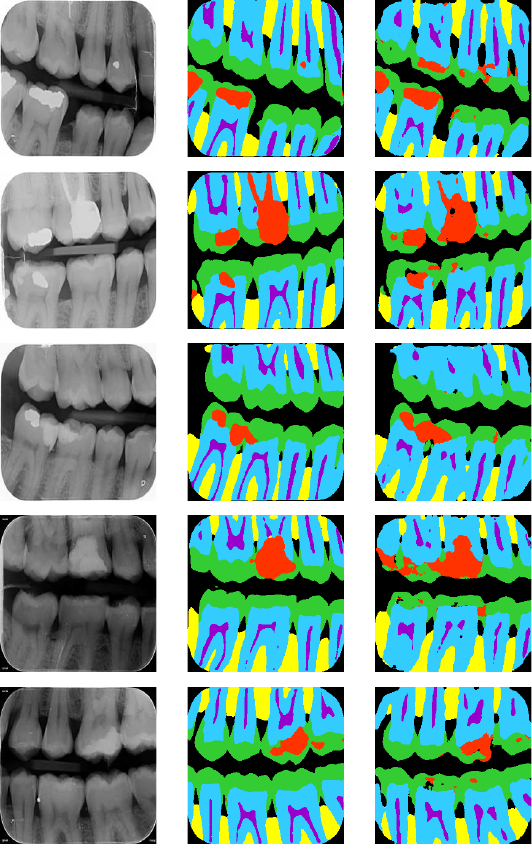}
    \caption{\textbf{Generated samples from PTDR and DDPM-MLP}}
\end{figure}

\end{document}